\documentclass[conference,9pt]{IEEEtran}

\usepackage{graphics}
\usepackage{booktabs}
\usepackage{graphicx}
\usepackage{subcaption}
\expandafter\def\csname ver@subfig.sty\endcsname{}
\usepackage{svg}
\usepackage{hyperref}

\usepackage{amsfonts}
\usepackage{amsmath}
\usepackage{multirow}
\usepackage{xcolor}
\usepackage{lipsum}
\usepackage{subfig}
\usepackage{xcolor}
\definecolor{mygreen}{RGB}{79,173,91}
\usepackage{algorithm}
\usepackage{algpseudocode}
\algtext*{EndWhile}
\algtext*{EndIf}
\algtext*{EndFor}
\algtext*{EndProcedure}
\usepackage[super]{nth}
\usepackage[normalem]{ulem}

\newcommand{\blue}[1]{\textcolor{blue}{#1}} 

\makeatletter
\def\@mkbibcitation{\relax}
\makeatother

\DeclareUnicodeCharacter{FF0C}{,}
\DeclareUnicodeCharacter{FF08}{(}
\DeclareUnicodeCharacter{FF09}{)}
\DeclareUnicodeCharacter{FF5E}{\textasciitilde} 

\begin{document}





\title{PatternPaint: Practical Layout Pattern Generation \\Using Diffusion-Based Inpainting}












\author{\IEEEauthorblockN{Guanglei Zhou\IEEEauthorrefmark{1},
Bhargav Korrapati\IEEEauthorrefmark{2},
Gaurav Rajavendra Reddy\IEEEauthorrefmark{2}, 
Chen-Chia Chang\IEEEauthorrefmark{1}, 
Jingyu Pan\IEEEauthorrefmark{1}, \\
Jiang	Hu\IEEEauthorrefmark{3},
Yiran Chen\IEEEauthorrefmark{1}, and
Dipto G. Thakurta\IEEEauthorrefmark{2}
}
\IEEEauthorblockA{\IEEEauthorrefmark{1}Dept. of Electrical \& Computer Engineering, Duke University， Durham, USA}
\IEEEauthorblockA{\IEEEauthorrefmark{2}Intel Corporation, Hillsboro, USA}
\IEEEauthorblockA{\IEEEauthorrefmark{3}Dept. of Electrical \& Computer Engineering, TAMU， College Station, USA}}

 
\maketitle 

\begin{abstract}

Generating diverse VLSI layout patterns is essential for various downstream tasks in design for manufacturing, as design rules continually evolve during the development of new technology nodes. 
However, existing training-based methods for layout pattern generation rely on large datasets. 
In practical scenarios, especially when developing a new technology node, obtaining such extensive layout data is challenging. Consequently, training models with large datasets becomes impractical, limiting the scalability and adaptability of prior approaches.

To this end, we propose PatternPaint, a diffusion-based framework capable of generating legal patterns with limited design-rule-compliant training samples. 
PatternPaint simplifies complex layout pattern generation into a series of inpainting processes with a template-based denoising scheme. 
Furthermore, we perform few-shot finetuning on a pretrained image foundation model with only 20 design-rule-compliant samples.
Experimental results show that using a sub-3nm technology node (Intel 18A), our model is the only one that can generate legal patterns in complex 2D metal interconnect design rule settings among all previous works and achieves a high diversity score. 
Additionally, our few-shot finetuning can boost the legality rate by 1.87X compared to the original pretrained model. As a result, we demonstrate a production-ready approach for layout pattern generation in developing new technology nodes.



\end{abstract}

\maketitle
 
\section{Introduction}



Generating synthetic pattern libraries is an essential and high-value element in technology development. 
However, this process faces significant challenges at advanced technology nodes. Engineers must first understand hundreds of design rules (DRs), and then create or modify pattern generators accordingly, resulting in lengthy turnaround times and substantial engineering effort.
Moreover, this becomes more challenging as the DRs are constantly changing at the early stage of technology development. Each new DR set requires diverse patterns to support many downstream tasks, such as optical proximity correction (OPC) elements ~\cite{OPC1,OPC2,OPC3,OPC4,OPC5,OPC_Pat_gen}, hotspot detection~\cite{Gaurav_Hotspot_detection,Hotspot1,Hotspot2,Hotspot3,Hotspot4}, design rule manual qualification~\cite{deesign_rule_manual}. 
These tasks require a wide spectrum of patterns to test/improve their methodologies and avoid unanticipated patterns that cause systematic failure. 

Before the rise of machine learning, several rule/heuristic-based methods~\cite{LiSPIE16,VIPER,VTS} were proposed to generate synthetic layout patterns. However, these heuristic methods demanded substantial engineering effort during development, as they required hundreds of design rules to be converted into algorithmic constraints. Moreover, these methods were often closely coupled with the DR set of a specific technology node, resulting in considerable time and effort to adjust them for new technology nodes. More recently, leveraging recent advances in generative models~\cite{EDALLM_survey,DiffusionbeatGAN}, a number of training-based ML methods such as GANs, Transformers, TCAEs, and Diffusion models~\cite{Deepattern, Diffpattern, LayouTransformer,CUP,ControLayout,wang2024chatpattern} have been proposed with the promise of reduced engineering effort and high pattern diversity. 

Despite these advancements, their practical application remains limited due to their dependence on large training datasets of clean DR layout samples. 
This limitation becomes particularly challenging in the early stage of technology node development. During this stage, design rules are continuously changing, and very few realistic layout patterns are available. Creating these thousands of training samples often requires rule-based methods to be coded as a pre-requisite, which demands significant time and effort. These constraints significantly restrict the deployment of training-based methods in critical Design for Manufacturability (DFM) applications.

Additionally, these works have been demonstrated only in oversimplified academic design rule settings, rather than being tested in close-to-realistic scenarios. 
Due to the simplicity of the rule set, they decompose layout generation into generating pattern topologies (a blueprint of a layout pattern consisting only of the shape of patterns) and using a nonlinear solver to convert the topology into DR-clean layout patterns. 
However, this decomposition becomes unrealistic due to the following reasons. First, the solver's runtime grows exponentially with both the number of designs and pattern size. Second, when the DR set includes discrete constraints, the problem transforms into a mixed integer programming problem, resulting in significantly lower legality rates using the original nonlinear setting.


\begin{figure}[t]
    \centering
    \includegraphics[width=1\linewidth]{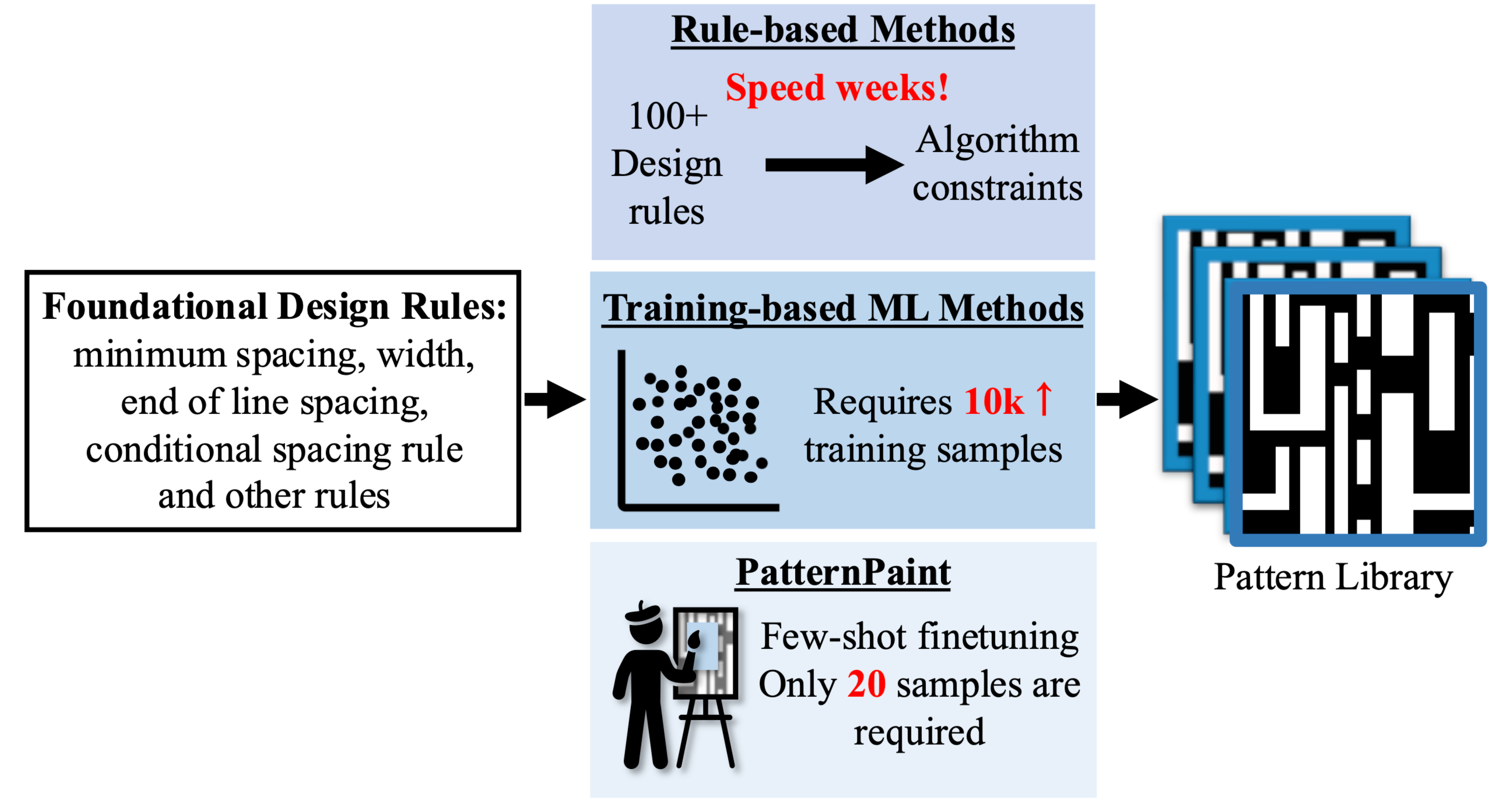}
    \caption{Comparison between rule-based methods, training-based methods~\cite{Diffpattern,wang2024chatpattern}, and our PatternPaint for layout pattern generation. 
    }
    \label{fig:header}
\end{figure}

To address these challenges, we propose PatternPaint, a few-shot inpainting framework capable of generating legal patterns. Our unique advantage is highlighted in Figure \ref{fig:header}. Our framework primarily targets single metal layer pattern generation to support DFM tasks, such as pattern feasibility analysis using OPC models.
PatternPaint simplifies layout generation into a series of inpainting processes, which naturally exploit design rule information encoded in neighboring regions of existing patterns. 
PatternPaint operates at the pixel level through a customized template-based denoising scheme, bypassing the need for solver-based legalization. 

Our contributions are outlined as follows:
\begin{itemize}
\item We present PatternPaint, the first few-shot pattern generation framework that leverages inpainting to drastically reduce training sample requirements for legal pattern generation. 

\item We decompose layout pattern generation into a series of inpainting processes with a novel template-based denoising scheme specifically designed for layout patterns. Our denoise method outperforms the conventional denoising method~\cite{opencv_nlm_2019} by achieving a tenfold increase in legal pattern generation.

\item \textbf{Validation on industrial PDKs:} PatternPaint is the first ML approach validated on an industrial PDK Intel 18A with full-set sign-off DR settings. 
Using only 20 starter samples, PatternPaint generates over 4000 DR-clean patterns, while prior ML solutions fail to deliver DR-clean patterns by training with 1k samples.

\end{itemize}

\section{Preliminaries}
    
\subsection{Diffusion model}

Diffusion models~\cite{DDPM} are generative models that operate through forward and reverse diffusion processes. The forward process gradually adds Gaussian noise to data over T timesteps:
\begin{align}
q(x_t|x_{t-1}) &= \mathcal{N} (x_t;\sqrt{1-\beta_t}x_{t-1};\beta_t \mathbf{I}) \\
q(x_1,...,x_T| x_0) &= \prod_{t=1}^T q(x_t|x_{t-1})
\end{align}
where $x_0$ is the original sample, $x_t$ represents noise-corrupted samples, and $\beta_t$ controls the noise schedule. As T increases, the data distribution approaches Gaussian:
\begin{align}
q(x_T) \approx \mathcal{N} (x_t;0;\mathbf{I})
\end{align}
The reverse process generates samples by learning to denoise:
\begin{align}
p_\theta (x_0) &= \int \prod_{t=1}^T p_\theta(x_{t-1}|x_t) dx_{1:T} \\
p_\theta(x_{t-1} | x_t) &= \mathcal{N} (x_{t-1};\mu_\theta(x_t,t);\sum_\theta(x_t,t))
\end{align}
where $\theta$ represents neural network parameters trained to minimize the objective:
\begin{align}
L &= \sum_{t=0}^T D_{KL}(q(x_t|x_{t+1},x_0) || p_\theta (x_t|x_{t+1})), t\in[1,T-1] 
\end{align}
For inpainting tasks~\cite{inpaint}, this process is conditioned on known image regions to fill masked areas consistently with the surrounding content. We found this characteristic aligns well with VLSI layout pattern generation, as design rule information is largely encoded in neighboring regions. By conditioning the generation on legal neighboring layouts, our model naturally learns to produce design-rule-compliant patterns. In later sections, we demonstrate how this approach significantly reduces training sample requirements while enriching pattern libraries.

\subsection{Squish representation}
\label{sec:squish}
\begin{figure}[t]
    \centering
    \includegraphics[width=0.9\linewidth]{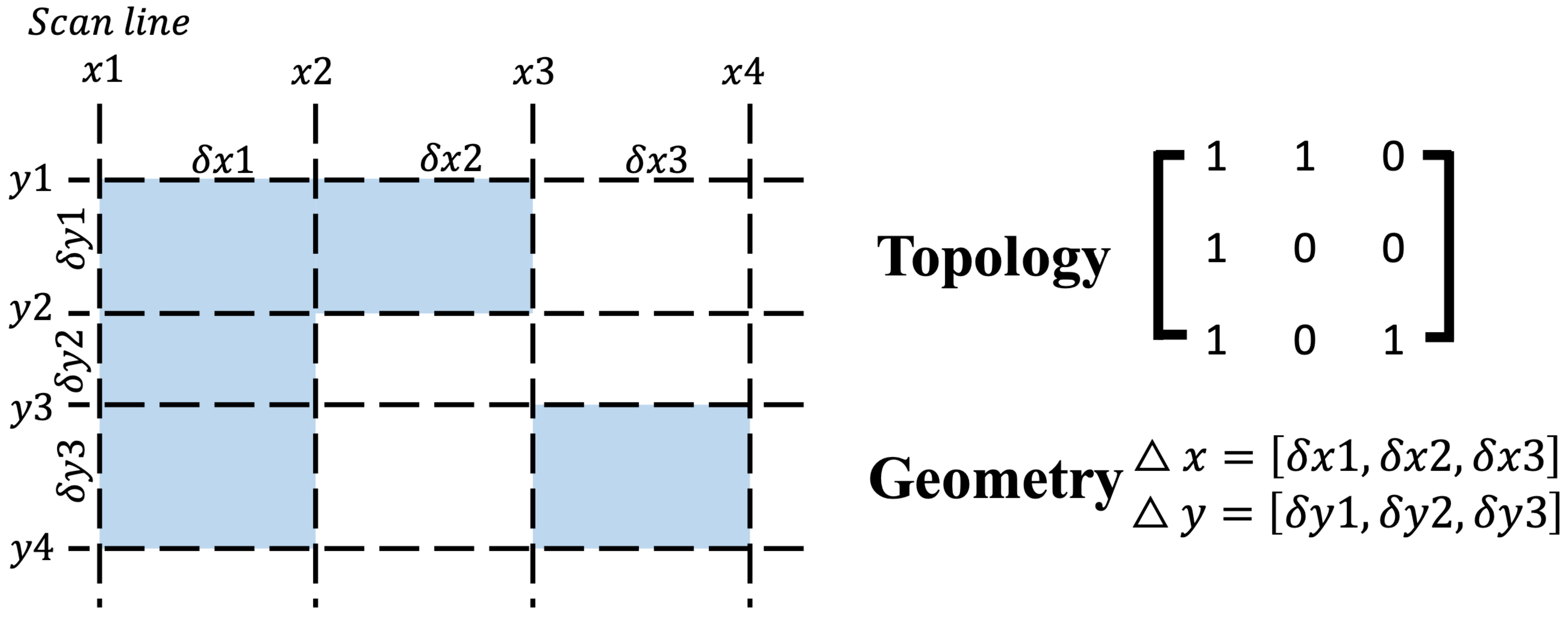}
    \caption{Squish Pattern Representation.}
    \label{fig:squish}
\end{figure} 

Standard layout patterns are typically composed of multiple polygons, presenting a sparse informational structure. To efficiently represent these patterns, majority of existing training-based methods~\cite{CUP, Deepattern, Diffpattern, CUP-EUV, wang2024chatpattern} use  ``Squish" pattern~\cite{squish,adapative_squish} to address these issues by compressing a layout into a concise pattern topology matrix alongside geometric data $(\triangle_x, 
 \triangle_y)$, as illustrated in Fig. \ref{fig:squish}. This process involves segmenting the layout into a grid framework using a series of scan lines that navigate along the edges of the polygon. The distances between each adjacent pair of scan lines are recorded in the $\triangle$ vectors. The topology matrix itself is binary, with each cell designated as either zero (indicating an absence of shape) or one (indicating the presence of a shape). 

Existing training-based methods focus on topology generation, leaving the geometry vector solved by a nonlinear solver. We refer to all the prior approaches that use squish representation as \textbf{squish-based} solutions. 
However, these solvers lack scalability and cannot handle advanced design rules effectively. 
To overcome these limitations, PatternPaint switches to a simpler \textbf{pixel-based} representation where $\triangle x_i, \triangle y_i$ are pre-defined with fixed physical widths (e.g., 1nm × 1nm rectangles per pixel).

\begin{figure}[t]
    \centering
    \includegraphics[width=0.49\textwidth]{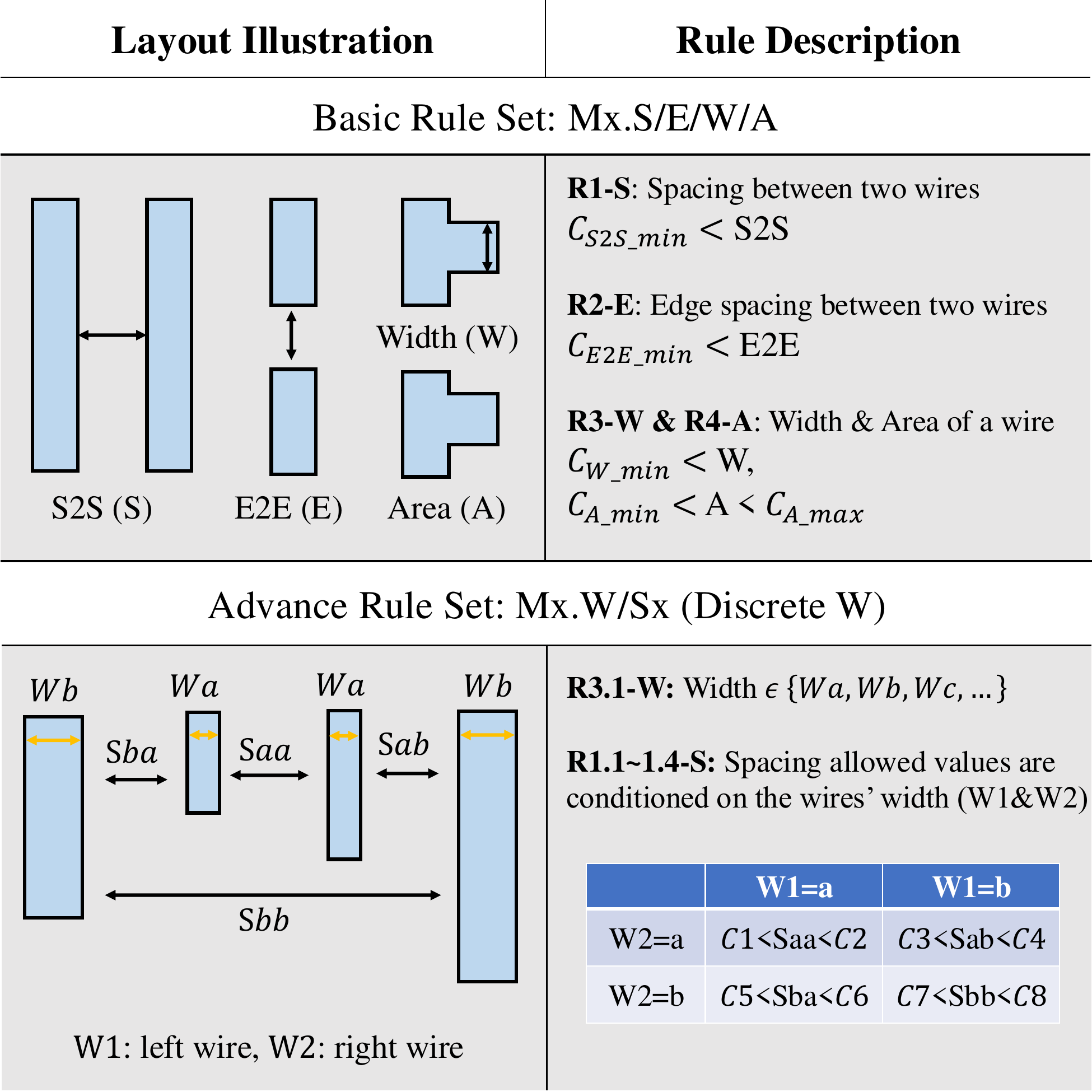}    \caption{Illustration of metal layer design rules. A selected set of design rules used in PatternPaint evaluation is shown as the advance rule set. (1) Basic Rule Set: Spacing (R1-S,R2-E)/ Width (R3-W)/ Area (R4-A) of Mx layer metal element. (2) Advance Rule Set: (R3.1-W) Only a set of discrete widths is allowed. (R1.1$~$1.4-S), the allowed spacing range is different depending on the neighboring metal widths.}
    \label{fig:DR}
\end{figure}


\subsection{Related Works}

Recent years, the development of training-based ML solutions for layout pattern generation has emerged. 
DeePattern~\cite{Deepattern} pioneered this field by using a Variational Autoencoder (VAE)~\cite{VAE_autoencoder} model to generate 1D layout patterns for 7nm EUV unidirectional settings with fixed metal tracks. 
This method employed a squish representation and a nonlinear solver to solve for the geometry vectors. 
Then, CUP~\cite{CUP} expanded the approach to 2D pattern generation, creating a large 2D academic layout pattern dataset containing 10k of training samples for a simple design rule setting (minimum width, spacing, and area). 
Under this dataset, LayouTransformer~\cite{LayouTransformer} introduced transformer-based sequential modeling, and DiffPattern~\cite{Diffpattern} employed discrete diffusion methods. Additional explorations include transferability~\cite{CUP-EUV}, free-size pattern generation~\cite{wang2024chatpattern}, and controllable generation~\cite{ControLayout}.

All existing ML works have been demonstrated only in basic DR settings. 
In contrast, PatternPaint addresses a full set of industrial standard DRs. 
As shown in Figure~\ref{fig:DR}, the advanced rule set introduces significantly more complex constraints. 
Under these complex rules, the nonlinear solver-based legalization used in existing state-of-the-art methods~\cite{Diffpattern,wang2024chatpattern} becomes unscalable due to the presence of discrete widths and upper bounds on spacing. 
This scalability issue is evident in~\cite{wang2024chatpattern}, where lower success rates are observed as pattern sizes increase. 
PatternPaint's pixel-based approach overcomes these limitations, enabling efficient pattern generation under more realistic and complex design rule constraints.
The limitation of the solver is further discussed in Section~\ref{sec:abalation}.

\section{Problem Formulation}
In this section, we formalize the pattern generation problem and its evaluation criteria. The objective is to produce diverse, realistic layout patterns from a small set of existing designs while ensuring compliance with rigorous design rules. We use a variety of metrics for quantitative evaluation.



(1) Legality: A layout pattern is legal iff it is DR-clean. 

(2) Entropy $H_1$:
As detailed in \cite{Diffpattern}, the complexity of a layout pattern is quantified as a tuple $(C_x,C_y)$ representing the count of scan lines along the x-axis and y-axis, respectively, each reduced by one. Then, we can obtain $$H_{1} = \sum_{i,j} P(C_{x_i},C_{y_j}) logP(C_{x_i},C_{y_j})$$ where $P(C_{x_i},C_{y_j})$ is the probability of encountering a pattern with complexities $C_{x_i}$ and $C_{y_j}$ within the library. 
This metric only focuses on topology diversity without considering any geometric information from actual patterns. 

(3) Entropy $H_2$:
To further examine the diversity of actual patterns with their geometric information included, we introduce $H_2$. For each unique combination of $\triangle x $ and $\triangle y$ (defined in Section~\ref{sec:squish}) presented in the library, we record their probability $P({\Delta_{x_i},\Delta_{y_j}})$ of having a pattern with the same $\Delta_{x,y}$ matrix within the library.
$$
H_{2} = \sum_{i,j} P({\Delta_{x_i},\Delta_{y_j}}) logP({\Delta_{x_i},\Delta_{y_j}})
$$
Since we target on pixel level generation, $H_2$ serves as the main metric to evaluate generation performance.


Based on the aforementioned evaluation metrics, the pattern generation problem can be formulated as follows. \\
\textbf{Problem 1 (Pattern Generation)}.  \textit{Given a set of design rules and existing patterns, the objective of pattern generation is to synthesize a legal pattern library such that $H_2$ of the layout patterns in the library is maximized.} 


\section{PatternPaint}
\label{sec:algorithm}
\begin{figure}[t]
    \centering
    \includegraphics[width=\linewidth]{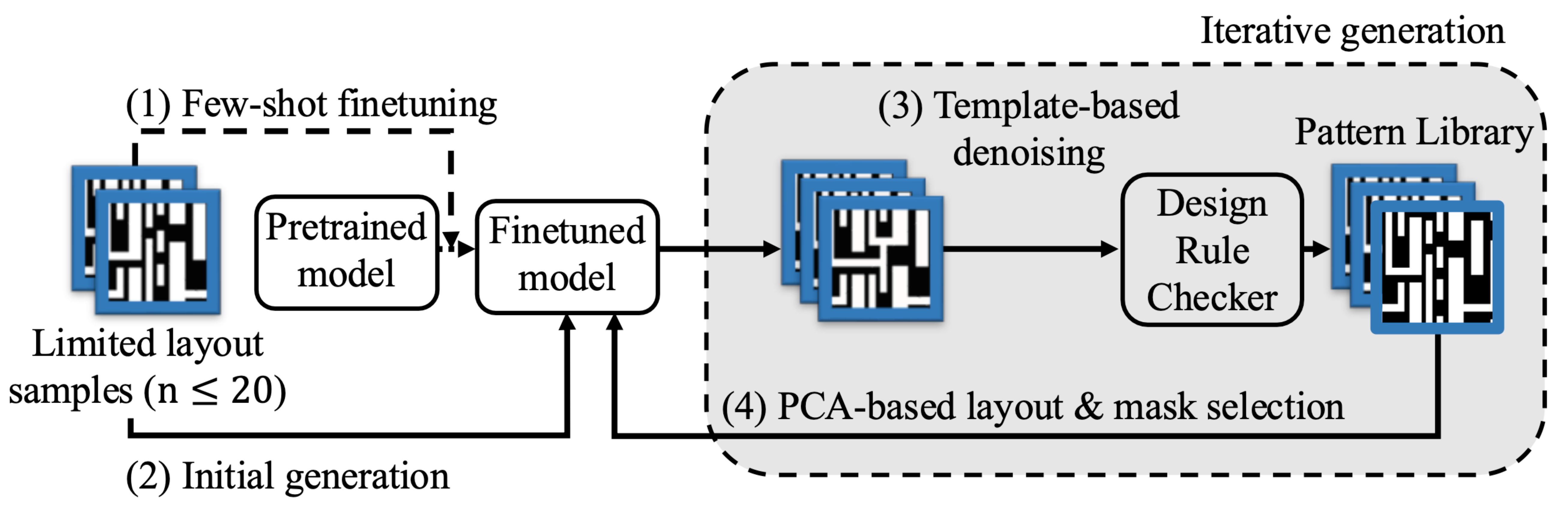}
    \caption{PatternPaint framework. It consists of: 
    (1) few-shot finetuning, (2) initial generation, (3) template-based denoising for layout refinement, followed by a design rule checking validation, and (4) PCA-based layout \& mask selection to select the next inpainting samples for iterative generation. This approach enables efficient pattern generation while ensuring design rule compliance. }
    \label{fig:framework}
\end{figure}

    

\subsection{Overview} 
As illustrated in Figure~\ref{fig:framework}, PatternPaint integrates four key components to achieve efficient layout pattern generation: (1) few-shot finetuning, (2) initial generation, (3) template-based denoising and (4) PCA-based layout \& mask selection. In the following sections, we detail how these components work together to produce diverse and legal layout patterns.


\subsection{Few-shot Finetuning}
PatternPaint adapts a pretrained text-to-image diffusion model to VLSI layouts through few-shot finetuning using the limited available layout samples. During finetuning, we fixed the text encoder and only finetuned on the image diffusion model. This finetuning shares a similar training objective with Equation (6) but starts from the $\theta'$ in the pretrained diffusion neural network. The training objectives then become
\begin{align}
L &= D_{KL}(q(x_T|x_0) || p_{\theta'} (x_T)) -logp_{\theta'}(x_0 | x_1) \\\nonumber 
&+\sum_{t=2}^{T} D_{KL}(q(x_t|x_{t+1},x_0) || p_{\theta'} (x_t|x_{t+1})) \\\nonumber 
&+ \lambda L_{prior}
\end{align}
where $x_0$ is selected from the limited $n$ layout patterns at the finetuning stage, $L_{prior}$ represents the prior preservation loss calculated on a set of class-specific images generated before training, and $\lambda$ is a weighting factor that balances the influence of prior preservation. The $L_{prior}$ helps serve as a regularization term to enable the model to learn very sparse samples while avoiding overfitting. The class-specific images are obtained by giving a fixed prompt to a text-to-image pretrained model. The interested reader is referred to \cite{ruiz2022dreambooth,stablediffusion} for details.

\subsection{Initial Pattern Generation}
After finetuning, PatternPaint begins the initial generation phase using the $n$ starter patterns from finetuning. Unlike prior approaches that generate entire patterns at once, our method decomposes generation into localized inpainting processes, mimicking how human engineers make targeted adjustments while preserving surrounding structures.

The generation process requires two inputs:
(1) starter patterns that provide design rule context and
(2) mask images that specify regions for variation.
A masked image $x_0^{masked}$ is created by applying the mask to a starter pattern, where masked regions (region replaced with Gaussian noise) are set to be predicted while unmasked regions remain unchanged. We provide 10 predefined masks (illustrated in Figure~\ref{fig:mask_selection}), though users can customize masks to target specific regions of interest. For each starter pattern-mask combination, the model generates multiple variations, producing a total of $n \times 10 \times v$ patterns in the initial iteration, where n is the number of starter patterns and v is the number of variations generated during inpainting.

\noindent \textbf{Inpainting}. During inpainting, the model predicts the masked regions while conditioning on the known pixels. The reverse diffusion process is modified as:

{\scriptsize
\begin{align}
    p_\theta(x_{t-1} | x_t, x_0^{masked}) = N(x_{t-1};\mu_\theta(x_t,x_0^{masked},t), \sum_\theta(x_t,x_0^{masked},t))
\end{align}}

The mean and covariance now also depend on the original masked image $x_0^{masked}$, conditioning the reverse process on the known pixels. We also follow the inference scheme mentioned in \cite{stablediffusion} that only generates masks with about 25\% region of its target image size. 


\begin{figure}[t]
    \centering
    \includegraphics[width=\linewidth]{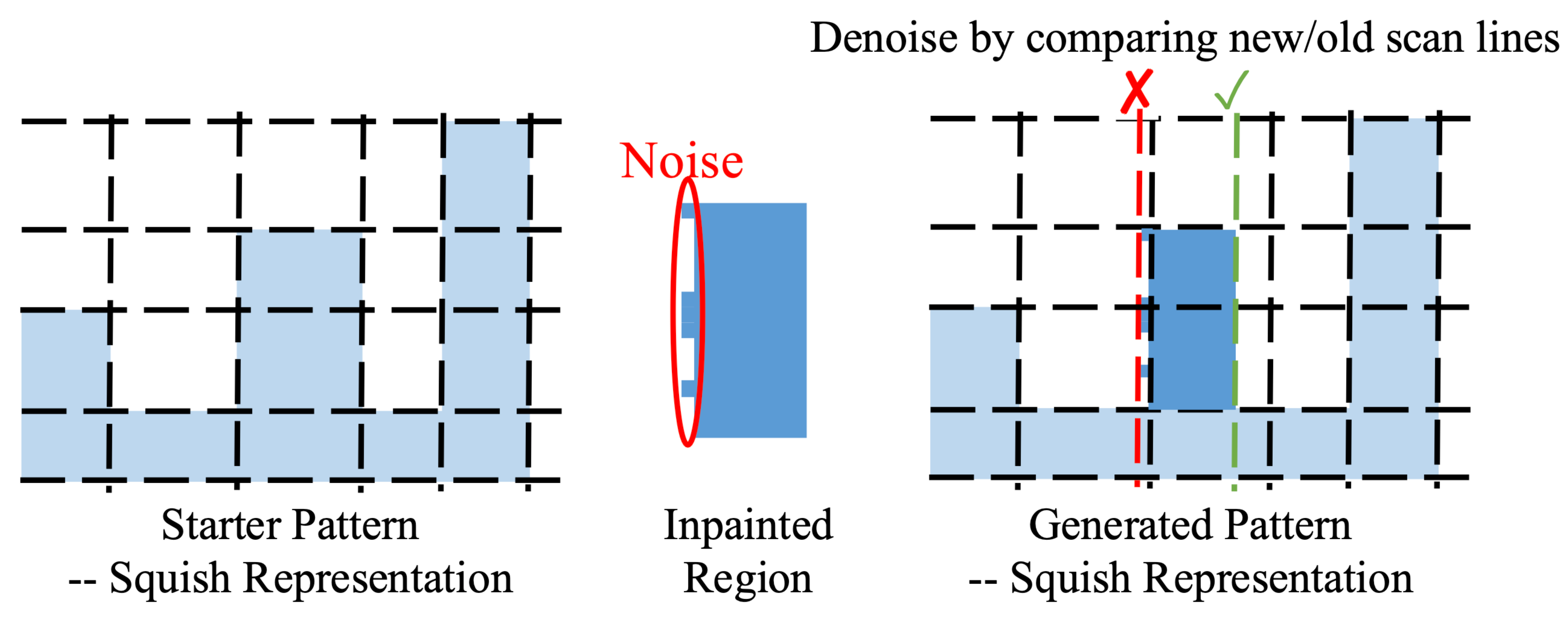}
    \vspace{-10pt}
    \caption{Illustration of Template-based Denoising. 
    Noise at the edge is reduced by comparing new scan lines with the original scan lines (black). Here, \textcolor{mygreen}{green} scan line is preserved since it is larger than a predefined threshold, and {\color{red} red} scan line is removed.}
    \label{fig:denoise}
\end{figure}

\begin{algorithm}
\caption{Template-based Denoising}
\label{alg:legalization}
\begin{algorithmic}[1]
\Require Generated noisy image \( I_g \), template (noise-free) image \( I_t \), and threshold \( T \)
\Ensure Denoised output image \( I_o \)
\State \( L_g \gets \textit{extract\_squish\_lines}(I_g) \)
\State \( L_t \gets \textit{extract\_squish\_lines}(I_t) \)
\State Cluster \( L_g \) into subsets \( C_1, C_2, \dots, C_n \) such that for each cluster \( C_i \), \(\| L_g(i) - L_g(j) \| \leq T \)
\For{each cluster \( C_i \)}
\State \( l_{\text{match}} \gets \) single scan line from \( L_t \) that minimizes \(\| l - C_i \|\)
    \If{\( \| l_{\text{match}} - C_i \| \leq T \)} 
        \State Replace \( C_i \) with \( l_{\text{match}} \) \Comment{Replace cluster with matched scan line from template}
    \Else
        \State Randomly select \( L_{\text{random}} \in C_i \) and replace the cluster with \( L_{\text{random}} \)
    \EndIf
\EndFor
\State Construct the topology matrix \( M \) from the modified \( L_g \)
\State \( I_o \gets \textit{reconstruct\_image}(M, L_g) \)\\
\Return \( I_o \)
\end{algorithmic}
\end{algorithm}

\subsection{Template-based Denoising and DR Checking}
The inpainting process, while effective for generating a big set of pattern variations, introduces noise along polygon edges due to the lossy nature of latent diffusion models. This edge noise can significantly alter pattern dimensions and lead to design rule violations. 

To address this challenge, we propose automated template-matching denoising, listed in Algorithm~\ref{alg:legalization}, inspired by the fact that only a sub-region of an image is changed during inpainting and the scan lines of the starter pattern (pre-inpainting) are known. We use the squish representation mentioned in Section 2.2, where we first extract scan lines from the noisy generated pattern (post-inpainting) and cluster similar lines within a predefined threshold. We then compare them to scan lines from the template (starter pattern). For each cluster, a parent scan line is chosen if available; otherwise, a line is randomly selected from within the cluster. This method is very effective, and we observe that it significantly increases the number of patterns passing DR checks. Figure~\ref{fig:denoise} also gives an intuitive example of denoising is performed by neglecting unnecessary scan lines due to edge noise but still preserving the scan lines. Denoising is performed by extracting the topology matrix using the designated scan lines and reconstructing the pattern again. A quantitative evaluation of this template-based denoising is shown in later Section~\ref{sec:abalation} and Table~\ref{tab:deonise_Comparison}.

\subsection{PCA-based layout \& mask selection}
\label{sec:iterative generation}

\begin{figure}[t]
    \centering
    \includegraphics[width=\linewidth]{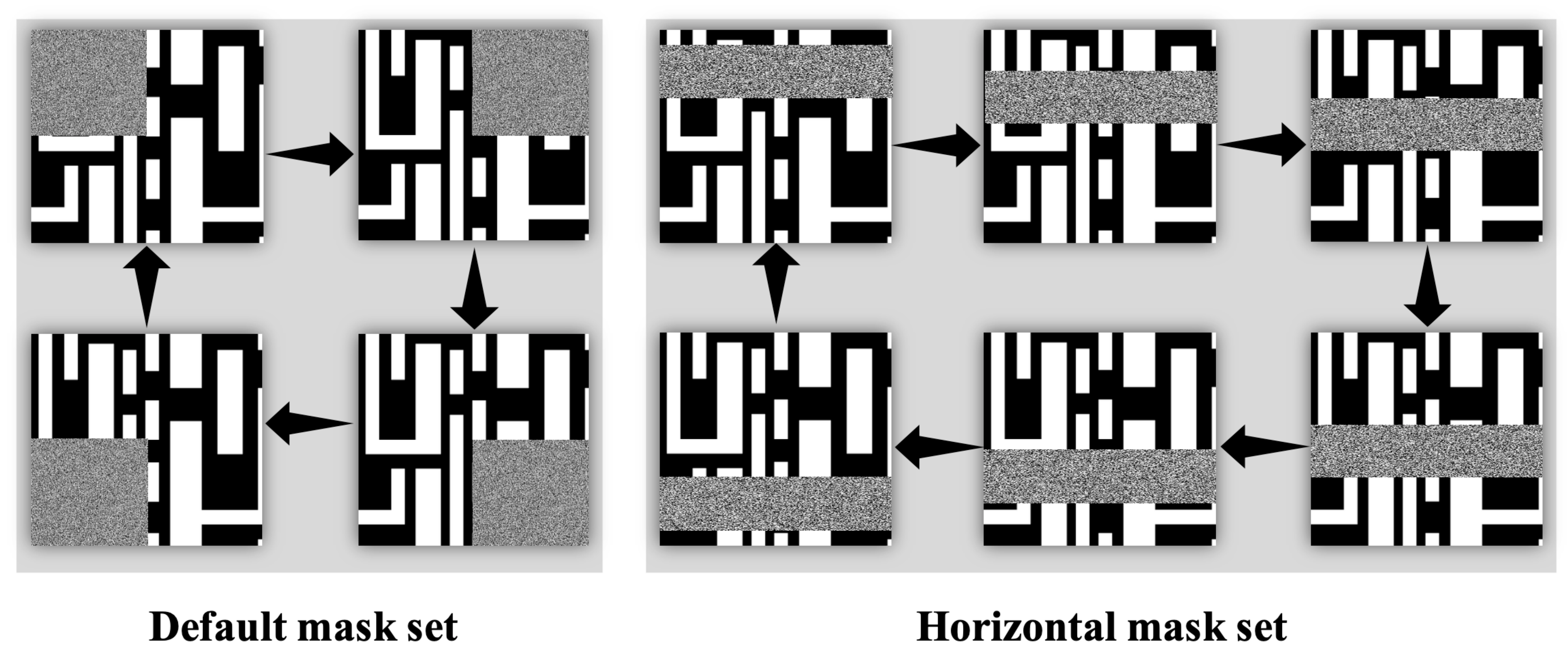}
    \caption{Predefined mask sets for pattern generation: default masks (left) and horizontal masks (right). Horizontal masks are customized for our dataset since we primarily focus on vertical track layout generation.  Mask in each set is selected sequentially during iterative generation. }
    \label{fig:mask_selection}
\end{figure}








\begin{algorithm}[t]
\caption{PCA-based Representative Layout Selection}
\begin{algorithmic}[1]
\Require Dataset $X \in \mathbb{R}^{n \times d}$, target samples $k$, constraints $C$
\Ensure Selected samples $S \subset X$

\State $X_{pca} \gets \text{PCA}(X)$ \Comment{Dimensionality reduction}
\State $I_s \gets \{\}$, $I_r \gets \{1,\ldots,n\}$ \Comment{Selected and remaining indices}

\State $i_0 \gets \text{random}(I_r)$ \Comment{Initial random sample}
\State $I_s \gets I_s \cup \{i_0\}$, $I_r \gets I_r \setminus \{i_0\}$

\For{$t \gets 1$ to $k-1$}
    \For{$i \in I_r$}
        \State $d_i \gets \sum_{s \in I_s} \|X_{pca}[i] - X_{pca}[s]\|$ \Comment{Sum of distances}
    \EndFor
    \State $i^* \gets \arg\max_{i \in I_r} d_i$ s.t. $C(X[i])$ \Comment{Farthest point}
    \State $I_s \gets I_s \cup \{i^*\}$, $I_r \gets I_r \setminus \{i^*\}$
\EndFor

\State \Return $X[I_s]$
\end{algorithmic}
\label{alg:sample}
\end{algorithm}

After the initial generation, a vast set of pattern variations is obtained. To produce more new and diverse layout patterns, iterative generation is employed, altering only a sub-region of the image in each iteration. For each iteration, we adopt a PCA-based approach to pick $k$ representative samples from the existing pattern library, followed by a mask selection scheme using two mask sets.

\subsubsection{PCA-based layout selection}

As described in Algorithm~\ref{alg:sample}, we propose a PCA-based layout selection scheme to pick representative layouts for the next iteration generation. PCA reduction provides a qualitative means to illustrate the diversity of a given layout pattern library \cite{VIPER}. The input samples are DR-clean layout clips. We first apply PCA to decompose images into several most representative components. 
To preserve most of the information in the dataset, we push the PCA to have explained\_varaince （0.9, meaning 90\% of the explained variance is preserved in the dimension-reduced components. 
Then, an iterative selection is performed to ensure that diverse samples are extracted from the existing library while meeting density constraints. The constraints can be easily integrated with other requirements such as specific pattern shapes or other interesting features and perform layout pattern generation in a more controlled setting.

\subsubsection{Mask selection}
As illustrated in Figure~\ref{fig:mask_selection}, our framework defined two mask sets (10 masks total) to guide pattern generation. The default mask set enables general pattern variations through targeted modifications, including metal wire modification and inter-track connections. The horizontal mask set is specifically designed for vertical track layouts to enhance exploration of end-to-end design rules and inner-track interactions. For horizontal track layout generation, a vertical mask set shall also be proposed. 

For each selected layout, we generate its mask following a predefined sequential schedule within each set. 
For example, when a pattern undergoes modification in one region (e.g., top-left in the default mask set), subsequent iterations target adjacent regions (e.g., top-right) following the predefined sequence. 
This sequential approach preserves features generated in previous iterations while providing rich contextual information for the inpainting model through newly generated patterns. 
\subsection{Iterative Generation}
As illustrated in the grey region of Figure~\ref{fig:framework}, the final iteration generation process then integrates Algorithm~\ref{alg:sample} to select representative layouts from the existing pattern library with a mask provided by its own mask set. Our framework keeps performing iterative generation until the desired diversity is reached or the sample budget is exceeded. When the iterations are completed, a diverse pattern library within the given DR space can be generated.

\section{Experimental Results}
\label{sec:experimental_result}

\subsection{Experimental Setup}
We validate PatternPaint on Intel 18A technology node with all generated patterns verified through industry-standard DR checking. 
The dataset contains 20 starter patterns. 


\noindent \textbf{Model setting}: 
We experiment on two pre-trained models, including stablediffusion1.5-inpaint (PatternPaint-sd1-base) and stablediffusion2-inpaint (PatternPaint-sd2-base)~\cite{stablediffusion}. 
 
\noindent \textbf{Finetuning details:} We adhered to the procedure described in DreamBooth \cite{ruiz2022dreambooth} to finetune the inpainting model with 20 layout patterns. 
The learning rate is set to 5e-6. 
For PatternPaint-sd1-base (PatternPaint-sd2-base), we denote its fine-tuned model as PatternPaint-sd1-ft (PatternPaint-sd2-ft). 
Experiments are performed on one Nvidia A100 GPU and one Intel(R) Xeon(R) Gold 6336Y CPU@2.40GHz.
Finetuning time takes around 10 minutes.
The average time for generating is 0.81 seconds and 0.21 seconds for denoising per sample.

\noindent \textbf{Baseline methods:}
We conducted comparisons using two state-of-the-art methods, CUP~\cite{CUP} and DiffPattern~\cite{Diffpattern}.
Since 20 patterns in our dataset are not enough to train diffusion-based and VCAE-based solution, we further obtain 1000 samples from a commercial tool with a size of 512 x 512 pixels to train CUP~\cite{CUP} and DiffPattern~\cite{Diffpattern} in squish representation~\cite{adapative_squish}.
The topology size for these experiments was set to 128 x 128 pixels. DiffPattern, which employs a non-linear solver-based legalization process, initially supported only three basic design rules. 
However, this is inadequate for Intel 18A, which includes constraints such as discrete values for certain line widths. 
We tried our best to improve this solver to accommodate a subset of the design rules that involve max-spacing, max-width, and discrete values for certain line widths. 
After this improvement, legal layout patterns started to appear. 
We implemented this nonlinear solver using scipy package, and the maximum iteration count is set to $10^8$.

\noindent \textbf{Initial generation:} 
Since previous works are one-time generation methods, to establish a fair comparison, we also present the results of the first stage of PatternPaint, initial generation, into performance comparison. 
For each initial pattern, each model generated 100 layout patterns per mask. In total, we generate 20,000 patterns. 
The performance of the initial generation across 4 models is denoted as (model-name)-init.

\noindent \textbf{Iterative generation:} 
Following the initial generation, we created a library of unique patterns with substantial variation. We then perform iterative pattern generation, as described in section~\ref{sec:iterative generation}, to check if diversity increased through this process. We designated the unique patterns from experiment 1 (Table 1) as our first iteration. For subsequent iterations, we conducted PCA analysis to select 100 of the most sparse representative samples, with the density constraint set at 40\% for the selected patterns. For each iteration, we generated 5000 samples out of the 100 patterns, adding only clean and new samples to our existing pattern library. We performed 6 iterative generations and collected 50000 generated patterns in total. The performance of iterative generation across 4 models is denoted as (model-name)-iter.


\subsection{Comparison of Pattern Generation}



\begin{table}[!tb]
\caption{Performance comparison for layout pattern generation.
 }
\label{tab:comparison-Intel4}
\scriptsize
\resizebox{\linewidth}{!}{
\begin{tabular}{lccccc}
\toprule
\textbf{Method} & \textbf{\begin{tabular}[c]{@{}c@{}}Generated\\ Patterns\end{tabular}}  & 
\textbf{\begin{tabular}[c]{@{}c@{}} Legal\\ Patterns\end{tabular}} &\textbf{\begin{tabular}[c]{@{}c@{}} Unique\\ Patterns\end{tabular}} & \textbf{$H_1$}& \textbf{$H_2$} \\ 
\midrule
 Starter patterns& - & 20 & 20 &3.68 & 4.32\\\\
CUP \cite{CUP} & 20000 & 0 & 0 & 0 & 0\\\\
 DiffPattern \cite{Diffpattern} & 20000 & 4 & 4 & 2 & 2 \\\\

PatternPaint-sd1-base-init &20000 & 1251 & 928 & 5.06& 9.78\\ \\
PatternPaint-sd2-base-init & 20000 & 1479 & 861& 5.15& 9.60\\\\
PatternPaint-sd1-ft-init & 20000 & 2336 & 1728 &4.65&  10.49 \\\\
PatternPaint-sd2-ft-init & 20000 & 1630 & 1469& 4.96 & 10.46 \\   \\

PatternPaint-sd1-base-iter & 50000 & 5021 & 3066 & 4.31 &11.37  \\ \\    
PatternPaint-sd2-base-iter & 50000 &5083  &2583 &4.19 & 11.02 \\ \\    
PatternPaint-sd1-ft-iter & 50000 & 7229 & 4458 & 4.08 &11.80  \\ \\    
PatternPaint-sd2-ft-iter & 50000 &5982  &4616 &4.11 & \textbf{12.01} \\     

\bottomrule
\end{tabular}
}
\end{table}

\begin{table}[!t]
\caption{Runtime comparison with our method and DiffPattern. 
 }
\label{tab:comparison-Intel4-runtime}
\scriptsize
\centering
\resizebox{0.6\linewidth}{!}{
\begin{tabular}{lc}
\toprule
\textbf{Method} & Avg Runtime (s) \\
\midrule
PatternPaint (Inpainting) & 0.81 \\
PatternPaint (Denoising) & 0.21 \\
DiffPattern & 38.04 \\
\bottomrule
\end{tabular}
}
\end{table}

\begin{figure*}
    \centering
    \includegraphics[width=\textwidth]{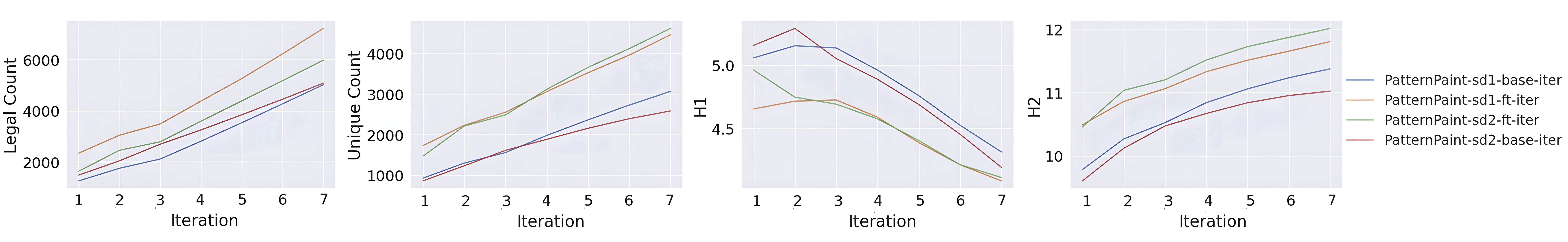}
    \caption{
    Experimental results for iterative generation process using four metrics: legal pattern counts, unique pattern counts, $H_1$, and $H_2$.
    }
    \label{fig:iterative}
\end{figure*}

The evaluation results of the initial generation are shown in Table~\ref{tab:comparison-Intel4}. 
10\% of the layout patterns generated by PatternPaint are legal and show better $H_1$ and $H_2$ than with the starter patterns.
Compared with other baselines, CUP is unable to generate legal patterns, and DiffPattern only generated four legal patterns.

The effectiveness of the proposed fine-tuning process is evident when comparing PatternPaint-sd1-base-init (and PatternPaint-sd2-base-init) with their fine-tuned counterparts, PatternPaint-sd1-ft-init (and PatternPaint-sd2-ft-init). Fine-tuned models show improvements in the number of legal patterns, unique patterns, and the main metric $H_2$. 
The results of iterative generation are in Figure~\ref{fig:iterative}.
As iterations proceed, both the unique pattern count and $H_2$ increase, further highlighting the gap between baseline models (PatternPaint-sd1-base, PatternPaint-sd2-base) and fine-tuned models (PatternPaint-sd1-ft, PatternPaint-sd2-ft), with the latter consistently outperforming the former. 
This validates that our fine-tuning process demonstrates significant model improvements.


We observe a slight decrease in $H_1$ as the iterative process proceeds, which can be attributed to the fact that $H_1$ primarily focuses on topology diversity. Since our framework alters only a sub-region of a given layout at a time, this results in several replicated topologies with adjustments limited to physical width, leading to the observed decrease in $H_1$. However, many DFM studies, such as OPC recipe development, benefit not only from topology diversity but also from variations in physical width combined with a given topology. This is captured by our key metric $H_2$, which considers both topology and physical dimensions. As iterations progress, more patterns with higher diversity are generated, including variations in physical widths and connection types. These diverse patterns can be used in yield analysis of metal patterns as well as finetuning OPC recipes, addressing the critical needs for real-world DFM applications.

\begin{figure}[t]
    \centering
    \begin{subfigure}[b]{0.15\textwidth}
    \includegraphics[width=\textwidth]{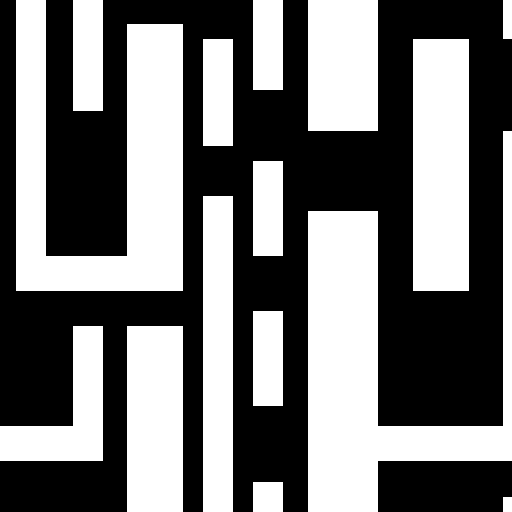}
    \caption{Starter pattern}
    \end{subfigure} 
    \begin{subfigure}[b]{0.15\textwidth}
    \includegraphics[width=\textwidth]{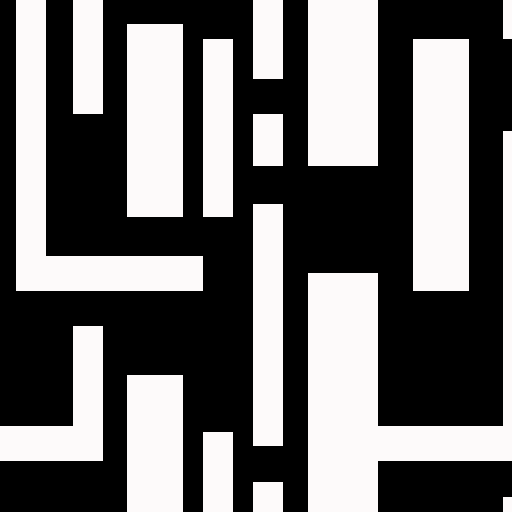}
    \caption{Generated pattern 1}
    \end{subfigure}
    \begin{subfigure}[b]{0.15\textwidth}
    \includegraphics[width=\textwidth]{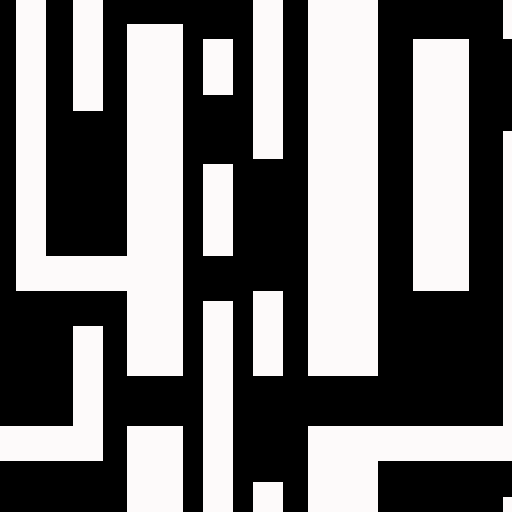}
    \caption{Generated pattern 2}
    \end{subfigure}
    \begin{subfigure}[b]{0.15\textwidth}
    \includegraphics[width=\textwidth]{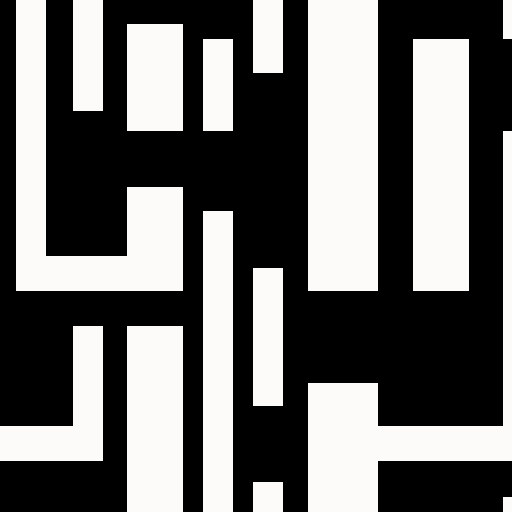}
    \caption{Generated pattern 3}
    \end{subfigure}
    \begin{subfigure}[b]{0.15\textwidth}
    \includegraphics[width=\textwidth]{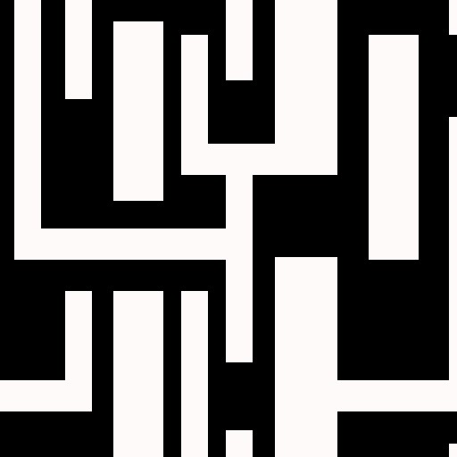}
    \caption{Generated pattern 4}
    \end{subfigure}
    \begin{subfigure}[b]{0.15\textwidth}
    \includegraphics[width=\textwidth]{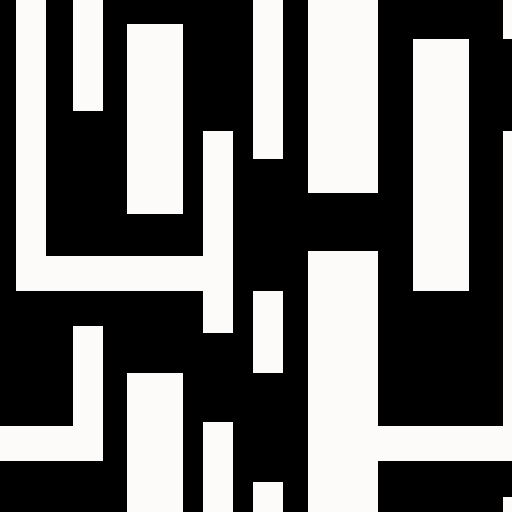}
    \caption{Generated pattern 5}
    \end{subfigure}
    \caption{Generated variations from a starter pattern.}
    \label{fig:variations}
\end{figure}

Figure~\ref{fig:variations} visually represents the variations generated by our proposed methods. The starter pattern is depicted in (a), (b-f) show the generated patterns. We observed that the proposed methods explored a wide range of variations, demonstrating the models' awareness of tracks. For example, in (f), the model attempts to disconnect from an adjacent thick track and establish a connection with a farther one. In (e), more complex changes were made, forming connections with even farther tracks and upper objects. These alterations enrich the pattern library and represent a unique feature of the ML-based method. Achieving such inter-track alternations with a rule-based method would require significant engineering effort, making it nearly impossible without advanced techniques coded for the given DR set.

Table~\ref{tab:comparison-Intel4-runtime} exhibits the runtime comparison with DiffPattern.
Note that we omit the comparison with CUP because it is unable to give a feasible solution.
The runtime for DiffPattern is 30$\times$ longer than PatternPaint due to the time-consuming solver-based legalization process.
Overall, these results demonstrate the effectiveness of PatternPaint to generate legal patterns and indicate that the solver-based solution is not suitable for industrial DR settings.

\section{Ablation Study}

We conduct comprehensive ablation studies to validate two key aspects: (1) the limitations of solver-based approaches with increasing design rule complexity, and (2) the effectiveness of our template-based denoising scheme.
\label{sec:abalation}
\begin{figure}[t]
    \centering
    \includegraphics[width=\linewidth]{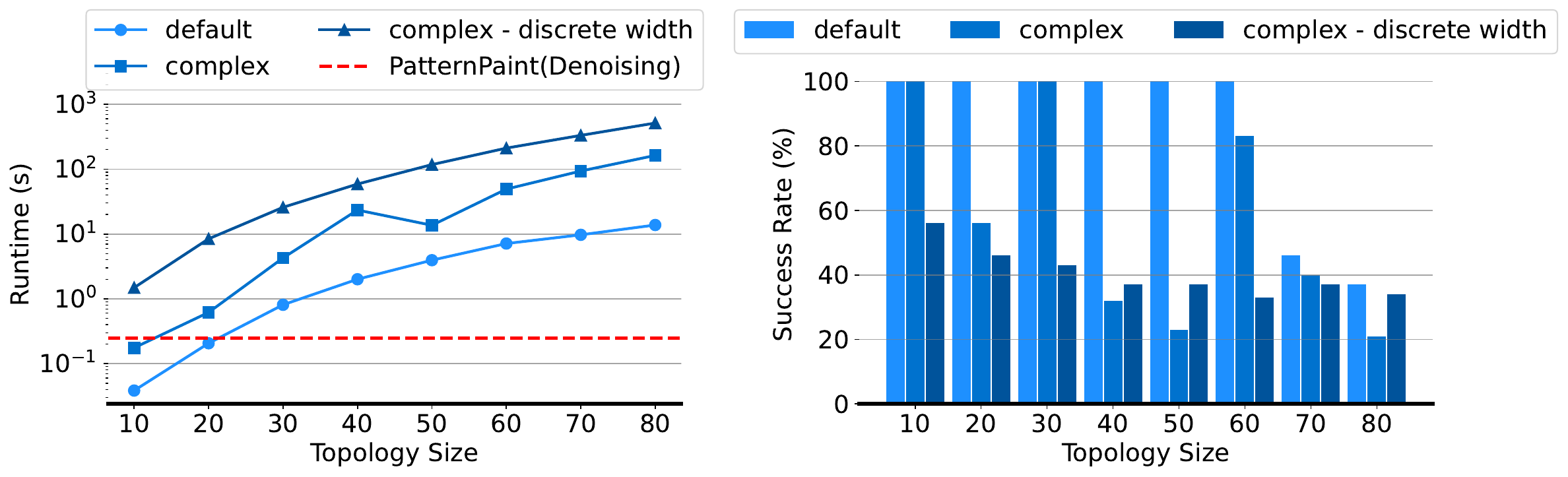}
    \caption{Runtime and success rate analysis of non-linear solver under three design rule settings: default, complex, and complex-discrete width. Results show exponential runtime growth and declining success rates with increasing topology size. 
    }
    \label{fig:solver-runtime}
\end{figure}

\subsubsection{Impact of Design Rule Complexity}
We evaluate three progressively design rule settings to illustrate the solver limitations as shown in Figure~\ref{fig:solver-runtime}.
The \textit{default} setting follows the academic design rule set of~\cite{Diffpattern}, including basic constraints such as minimum width/spacing and area checks. 
The \textit{complex} setting extends this by differentiating horizontal and vertical directions for width and spacing checks, including their minimum and maximum values. 
The \textit{complex-discrete} setting further restricts width values to discrete sets.
We observe two critical limitations of this non-linear solver. 
First, the solver's runtime increases significantly from default to complex-discrete settings, showing exponential growth with topology size, significantly exceeding PatternPaint's denoising time. 
Second, despite the existence of legal solutions, the solver's success rate deteriorates with rule complexity. 
For topologies larger than 60×60, all settings achieve less than 50\% success rate.
This scalability issue is also evident in~\cite{wang2024chatpattern}, where lower success rates are observed as pattern sizes increase. 

\subsubsection{Effectiveness of Template-based Denoising}

\begin{table}[t]
\caption{
Comparison of the pattern generation success rate (S\%) using PatternPaint with different denoising schemes: our template-based denoising, OpenCV non-local means filter~\cite{opencv_nlm_2019}, and without denoising.
}
 \label{tab:deonise_Comparison}
\scriptsize
\resizebox{\linewidth}{!}{
    \begin{tabular}{lccc}
        \toprule
\textbf{Method} & \textbf{\begin{tabular}[c]{@{}c@{}}W/ Template-\\based Denoise (S$\uparrow$)\end{tabular}}  & 
\textbf{\begin{tabular}[c]{@{}c@{}} W/ OpenCV\\ Denoise Filter \cite{opencv_nlm_2019} (S$\uparrow$)\end{tabular}} &\textbf{\begin{tabular}[c]{@{}c@{}} W/o\\ Denoise (S$\uparrow$)\end{tabular}}\\ 
        \midrule
        PatternPaint-sd1-base    & 6.25 & 0.12 & 0 \\ \\
        PatternPaint-sd1-ft    & 11.68 & 1.04 & 0 \\\\
        PatternPaint-sd2-base  & 7.40 &  0.24 & 0 \\\\
        PatternPaint-sd2-ft    & 8.15 & 0.76 & 0 \\
        \midrule
        \textbf{Average} & \textbf{8.37} & \textbf{0.86} & \textbf{0} \\
        \bottomrule
    \end{tabular}
}
\end{table}

Table~\ref{tab:deonise_Comparison} evaluates the effectiveness of our template-based denoising scheme in the PatternPaint framework. We compare the template-based denoising scheme with a widely used denoising filter, the non-local means filter~\cite{opencv_nlm_2019} implemented in OpenCV. We also show the DRC results without any denoising activity. The generation success rate is calculated by legal patterns divided by total generated patterns. 
The results show that no patterns can be directly used without denoise. Our template-based denoising significantly outperforms the OpenCV non-local means filter~\cite{opencv_nlm_2019}, with an average of 9.7x generation success rate improvement. The fine-tuned versions of PatternPaint achieve the highest success rates when combined with template-based denoising, reaching 11.68\%. 
These findings validate the effectiveness of the template-based denoising scheme in maximizing pattern generation efficiency, especially when combined with fine-tuning techniques.

\section{Conclusion}
\label{sec:conclusion}
In this paper, we propose PatternPaint, an automated few-shot pattern generation framework using diffusion-based inpainting. We develop our own unique template-based denoising scheme to tackle noise and propose a PCA-based sample selection scheme for iterative pattern generation. In the initial round of the iterative generation process, thousands of DR clean layouts are generated on the latest Intel PDK and checked through an industry-standard DR checker. In later iterations, by measuring entropy, we observed that pattern diversity improves. Our work, PatternPaint, has its unique benefits with little to no human effort in loop and is the first can perform pattern generation in a few-shot learning scenario. 
In future work, we will improve PatternPaint to support larger size pattern generation and explore further finetuning the pre-trained models using legal samples collected from PatternPaint enriched pattern library. We also plan to evaluate the explored design rule space against product-level layout patterns and demonstrate the application of PatternPaint-generated patterns on yield learning test chips for future PDK development and DFM studies on Intel silicon.


\section{Acknowledgement}
This work is partially supported by SRC 3104.001 and NSF 2106828.

\bibliographystyle{ieeetr}
\bibliography{reference}

\begin{thebibliography}{10}

\bibitem{OPC1}
J.-R. Gao, X.~Xu, B.~Yu, and D.~Z. Pan, ``{MOSAIC}: Mask optimizing solution with process window aware inverse correction,'' in {\em Proceedings of the Design Automation Conference (DAC)}, pp.~1--6, 2014.

\bibitem{OPC2}
Y.~Jiang, F.~Yang, B.~Yu, D.~Zhou, and X.~Zeng, ``Efficient layout hotspot detection via neural architecture search,'' {\em ACM Trans. Des. Autom. Electron. Syst.}, vol.~27, June 2022.

\bibitem{OPC3}
B.~Jiang, H.~Zhang, J.~Yang, and E.~F.~Y. Young, ``A fast machine learning-based mask printability predictor for opc acceleration,'' in {\em Proceedings of the 24th Asia and South Pacific Design Automation Conference (ASPDAC)}, pp.~412--419, 2019.

\bibitem{OPC4}
J.~Kuang, W.-K. Chow, and E.~F.~Y. Young, ``A robust approach for process variation aware mask optimization,'' in {\em Proceedings of the Design, Automation \& Test in Europe Conference \& Exhibition (DATE)}, pp.~1591--1594, 2015.

\bibitem{OPC5}
H.~Yang, S.~Li, Y.~Ma, B.~Yu, and E.~F.~Y. Young, ``{GAN-OPC}: Mask optimization with lithography-guided generative adversarial nets,'' in {\em Proceedings of the Design Automation Conference (DAC)}, pp.~1--6, 2018.

\bibitem{OPC_Pat_gen}
A.~Hamouda, M.~Bahnas, D.~Schumacher, I.~Graur, A.~Chen, K.~Madkour, H.~Ali, J.~Meiring, N.~Lafferty, and C.~McGinty, ``Enhanced opc recipe coverage and early hotspot detection through automated layout generation and analysis,'' in {\em Optical Microlithography XXX}, vol.~10147, p.~101470R, SPIE, 2017.

\bibitem{Gaurav_Hotspot_detection}
G.~R. Reddy, K.~Madkour, and Y.~Makris, ``Machine learning-based hotspot detection: Fallacies, pitfalls and marching orders,'' in {\em Proceedings of IEEE/ACM International Conference on Computer-Aided Design (ICCAD)}, pp.~1--8, 2019.

\bibitem{Hotspot1}
R.~Chen, W.~Zhong, H.~Yang, H.~Geng, X.~Zeng, and B.~Yu, ``Faster region-based hotspot detection,'' in {\em Proceedings of the Design Automation Conference (DAC)}, pp.~1--6, 2019.

\bibitem{Hotspot2}
J.~Pan, C.-C. Chang, Z.~Xie, J.~Hu, and Y.~Chen, ``Robustify ml-based lithography hotspot detectors,'' in {\em Proceedings of the IEEE/ACM International Conference on Computer-Aided Design (ICCAD)}, pp.~1--7, 2022.

\bibitem{Hotspot3}
H.~Zhang, B.~Yu, and E.~F. Young, ``Enabling online learning in lithography hotspot detection with information-theoretic feature optimization,'' in {\em Proceedings of IEEE/ACM International Conference on Computer-Aided Design (ICCAD)}, pp.~1--8, 2016.

\bibitem{Hotspot4}
H.~Yang, Y.~Lin, B.~Yu, and E.~F.~Y. Young, ``Lithography hotspot detection: From shallow to deep learning,'' in {\em Proceedings of the 30th IEEE International System-on-Chip Conference (SOCC)}, pp.~233--238, 2017.

\bibitem{deesign_rule_manual}
A.~Kabeel, S.~Kim, Y.~G. Park, D.~Kim, J.~Kwan, S.~Rizk, K.~Madkour, M.~Shafee, and J.~Kim, ``Design rule manual and drc code qualification flows empowered by high coverage synthetic layouts generation,'' in {\em DTCO and Computational Patterning II}, vol.~12495, pp.~415--428, SPIE, 2023.

\bibitem{LiSPIE16}
H.~Li, E.~Zou, R.~Lee, S.~Hong, S.~Liu, J.~Wang, C.~Du, R.~Zhang, K.~Madkour, H.~Ali, {\em et~al.}, ``Design space exploration for early identification of yield limiting patterns,'' in {\em Design-Process-Technology Co-optimization for Manufacturability X}, vol.~9781, 2023.

\bibitem{VIPER}
G.~R. Reddy, M.-M. Bidmeshki, and Y.~Makris, ``{VIPER}: A versatile and intuitive pattern generator for early design space exploration,'' in {\em Proceedings of the IEEE International Test Conference (ITC)}, pp.~1--7, 2019.

\bibitem{VTS}
G.~R. Reddy, C.~Xanthopoulos, and Y.~Makris, ``Enhanced hotspot detection through synthetic pattern generation and design of experiments,'' in {\em Proceedings of IEEE VLSI Test Symposium (VTS)}, pp.~1--6, 2018.

\bibitem{EDALLM_survey}
J.~Pan, G.~Zhou, C.-C. Chang, I.~Jacobson, J.~Hu, and Y.~Chen, ``A survey of research in large language models for electronic design automation,'' {\em ACM Trans. Des. Autom. Electron. Syst.}, vol.~30, Feb. 2025.

\bibitem{DiffusionbeatGAN}
P.~Dhariwal and A.~Nichol, ``Diffusion models beat gans on image synthesis,'' in {\em Proceedings of the International Conference on Neural Information Processing Systems (NeurIPS)}, 2021.

\bibitem{Deepattern}
H.~Yang, S.~Li, W.~Chen, P.~Pathak, F.~Gennari, Y.-C. Lai, and B.~Yu, ``Deepattern: Layout pattern generation with transforming convolutional auto-encoder,'' {\em IEEE Transactions on Semiconductor Manufacturing}, vol.~35, no.~1, pp.~67--77, 2022.

\bibitem{Diffpattern}
Z.~Wang, Y.~Shen, W.~Zhao, Y.~Bai, G.~Chen, F.~Farnia, and B.~Yu, ``Diffpattern: Layout pattern generation via discrete diffusion,'' in {\em Proceedings of the Design Automation Conference (DAC)}, pp.~1--6, 2023.

\bibitem{LayouTransformer}
L.~Wen, Y.~Zhu, L.~Ye, G.~Chen, B.~Yu, J.~Liu, and C.~Xu, ``{LayouTransformer}: Generating layout patterns with transformer via sequential pattern modeling,'' in {\em Proceedings of the IEEE/ACM International Conference on Computer-Aided Design (ICCAD)}, pp.~1--9, 2022.

\bibitem{CUP}
X.~Zhang, J.~Shiely, and E.~F. Young, ``Layout pattern generation and legalization with generative learning models,'' in {\em Proceedings of the IEEE/ACM International Conference On Computer Aided Design (ICCAD)}, pp.~1--9, 2020.

\bibitem{ControLayout}
Q.~Wang, X.~Zhang, M.~D. Wong, and E.~F. Young, ``Controlayout: Conditional diffusion for style-controllable and violation-fixable layout pattern generation,'' in {\em Proceedings of the Great Lakes Symposium on VLSI (GLSVLS)}, p.~511–515, 2024.

\bibitem{wang2024chatpattern}
Z.~Wang, Y.~Shen, X.~Yao, W.~Zhao, Y.~Bai, F.~Farnia, and B.~Yu, ``{ChatPattern}: Layout pattern customization via natural language,'' in {\em Proceedings of the Design Automation Conference (DAC)}, pp.~1--6, 2024.

\bibitem{opencv_nlm_2019}
``Image denoising.'' \url{https://docs.opencv.org/3.4/d5/d69/tutorial_py_non_local_means.html}, 2019.
\newblock Accessed: 2024-09-28.

\bibitem{DDPM}
J.~Ho, A.~Jain, and P.~Abbeel, ``Denoising diffusion probabilistic models,'' in {\em Proceedings of the International Conference on Neural Information Processing Systems (NeurIPS)}, pp.~6840--6851, 2020.

\bibitem{inpaint}
A.~Lugmayr, M.~Danelljan, A.~Romero, F.~Yu, R.~Timofte, and L.~Van~Gool, ``Repaint: Inpainting using denoising diffusion probabilistic models,'' in {\em Proceedings of the IEEE/CVF Conference on Computer Vision and Pattern Recognition (CVPR)}, pp.~11451--11461, 2022.

\bibitem{CUP-EUV}
X.~Zhang, H.~Yang, and E.~F. Young, ``Attentional transfer is all you need: Technology-aware layout pattern generation,'' in {\em Proceedings of the Design Automation Conference (DAC)}, pp.~169--174, 2021.

\bibitem{squish}
F.~E. Gennari and Y.-C. Lai, ``Topology design using squish patterns,'' {\em U.S. Patent 8832621B1}, 2014.

\bibitem{adapative_squish}
H.~Yang, P.~Pathak, F.~Gennari, Y.-C. Lai, and B.~Yu, ``Detecting multi-layer layout hotspots with adaptive squish patterns,'' in {\em Proceedings of the Asia and South Pacific Design Automation Conference (ASPDAC)}, p.~299–304, 2019.

\bibitem{VAE_autoencoder}
D.~P. Kingma and M.~Welling, ``Auto-encoding variational bayes,'' 2022.

\bibitem{ruiz2022dreambooth}
N.~Ruiz, Y.~Li, V.~Jampani, Y.~Pritch, M.~Rubinstein, and K.~Aberman, ``Dreambooth: Fine tuning text-to-image diffusion models for subject-driven generation,'' in {\em Proceedings of the IEEE/CVF Conference on Computer Vision and Pattern Recognition (CVPR)}, pp.~22500--22510, 2023.

\bibitem{stablediffusion}
R.~Rombach, A.~Blattmann, D.~Lorenz, P.~Esser, and B.~Ommer, ``High-resolution image synthesis with latent diffusion models,'' in {\em Proceedings of the IEEE/CVF Conference on Computer Vision and Pattern Recognition (CVPR)}, pp.~10684--10695, 2022.

\end{thebibliography}

\end{document}